\pdfoutput=1

\documentclass[11pt]{article}

\usepackage[final]{acl}

\usepackage{times}
\usepackage{latexsym}
\usepackage{graphicx}
\usepackage{amsmath} 
\usepackage{enumitem}
\usepackage{amsfonts}
\usepackage{booktabs}

\usepackage[T1]{fontenc}

\usepackage[utf8]{inputenc}

\usepackage{microtype}

\usepackage{inconsolata}

\usepackage{graphicx}

\title{Lacuna Inc. at SemEval-2025 Task 4: LoRA-Enhanced Influence-Based Unlearning for LLMs}

\usepackage{inconsolata}
\usepackage{hyperref}
\usepackage{fontawesome5}

\author{Aleksey Kudelya \textsuperscript{\faCrow}, 
        Alexander Shirnin\textsuperscript{\faCrow} \\
    \textsuperscript{\faCrow}HSE University \\
    \small{
    \textbf{Correspondence:} \href{mailto:ashirnin@hse.ru}{\texttt{ashirnin@hse.ru}}
}
}

\begin{document}
\maketitle
\begin{abstract}
This paper describes LIBU (LoRA enhanced influence-based unlearning), an algorithm to solve the task of unlearning - removing specific knowledge from a large language model without retraining from scratch and compromising its overall utility (SemEval-2025 Task 4: Unlearning sensitive content from Large Language Models). 
The algorithm combines classical \textit{influence functions} to remove the influence of the data from the model and \textit{second-order optimization} to stabilize the overall utility. Our experiments show that this lightweight approach is well applicable for unlearning LLMs in different kinds of task. 
\end{abstract}

\section{Introduction}

Machine unlearning - the process of removing specific knowledge from a machine learning model without retraining from scratch - has emerged as a critical capability for large language models (LLMs)~\cite{10.1007/978-3-031-61000-4_10} deployed in dynamic or privacy-sensitive environments \cite{8b63c9b621a148088f2f835e46fc1a61}. Unlike traditional retraining, which is computationally prohibitive for LLMs, unlearning seeks to selectively erase influences of a \textit{forget dataset} while preserving good performance on a \textit{retain set}. This capability is essential for applications requiring compliance with data privacy regulations (e.g., GDPR "right to be forgotten"~\cite{MANTELERO2013229}) or rapid adaptation to evolving content policies.

Existing approaches, for example, gradient ascent~\cite{Tarun_2024}, often degrade general capabilities, as reflected in performance drops on benchmarks like MMLU~\cite{hendryckstest2021}, or demand computational resources that are impractical in real-world settings. To address these limitations, we propose a two-phase method called LoRA-enhanced influence-based unlearning (LIBU), which combines influence functions~\cite{10.5555/3305381.3305576} with the Sophia optimizer~\cite{liu2024sophia}. In Phase 1, LIBU computes parameter-wise updates using a Fisher Information approximation~\cite{Foster_Schoepf_Brintrup_2024} to minimize retain-set disruption. Phase 2 refines the model via second-order optimization, stabilizing training on noisy forget-set gradients. Our submission, evaluated on the OLMo-7B model~\cite{groeneveld-etal-2024-olmo}, achieves a regurgitation rate of 0.283 while maintaining an MMLU accuracy of 0.469, exceeding the competition threshold of 0.371.

\section{Background}
The SemEval-2025 Task 4: Unlearning Sensitive Content from LLMs \cite{ramakrishna2025semevaltask4} formalizes this challenge across three subtasks: (1) erasing long-form synthetic documents, (2) removing personally identifiable information (PII) from short biographies, and (3) unlearning real documents from the OLMo pretraining corpus. The task demands balancing two competing objectives: achieving high regurgitation and MIA scores on forget and retain sets, while preserving highest MMLU capability. 
To evaluate the submissions, \cite{ramakrishna2025lumellmunlearningmultitask} release a comprehensive new benchmark named LUME (LLM Unlearning with Multitask Evaluations). For each of the tasks, there are prompts for regurgitation and knowledge tests. The benchmark is split into forget and retain sets (in 1:1 ratio). Two model checkpoints (7B and 1B parameters) were also fine-tuned to memorize this dataset.

\begin{figure*}[!th]
  \centering
  \includegraphics[width=0.95\textwidth]{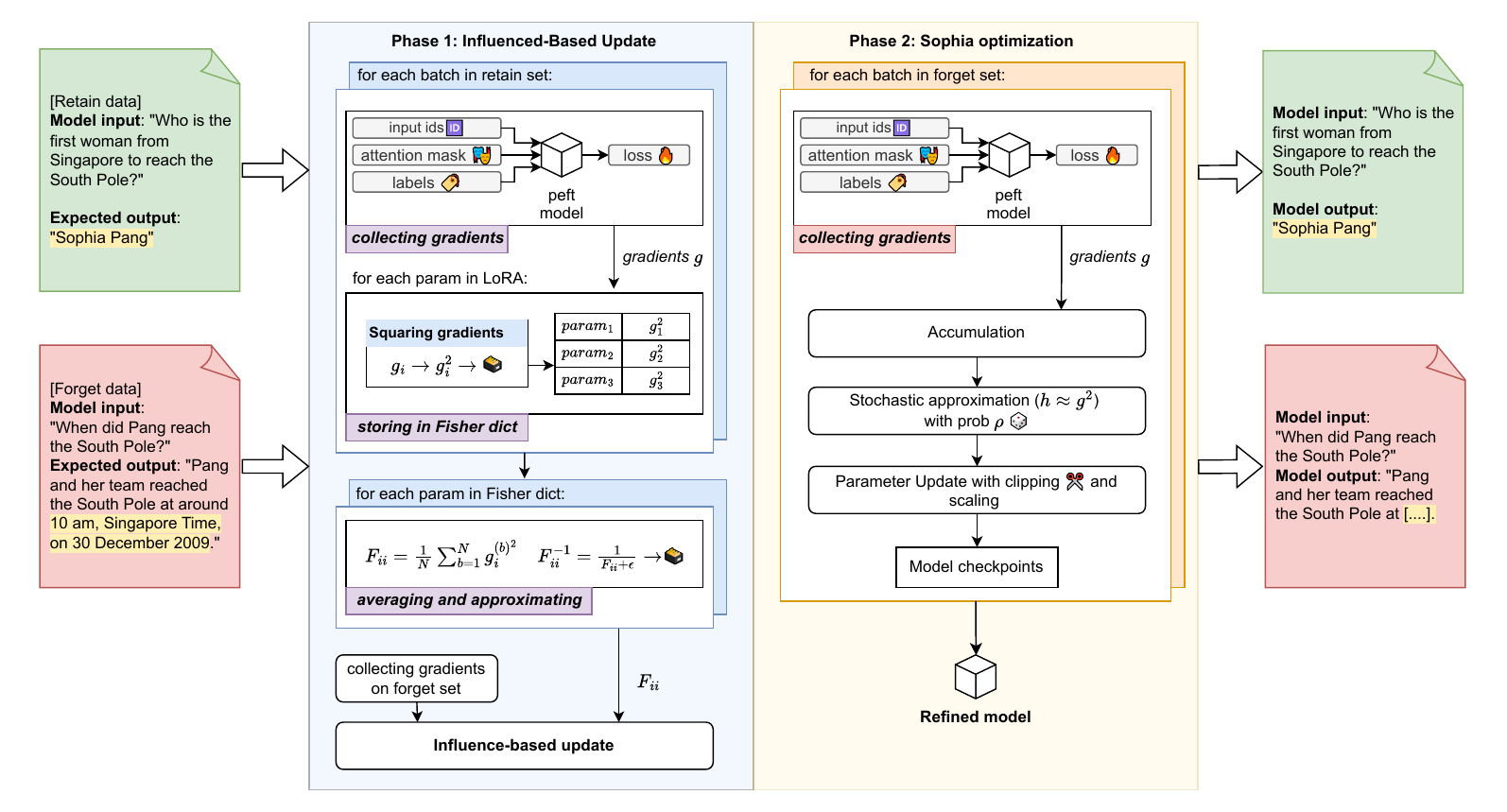}
  \caption{LIBU pipeline. Given two datasets, LIBU operates with two phases: \textbf{1) Influence-Based Update}, where it collects the gradients from retain and forget sets and determines the necessary parameter updates; \textbf{2) Sophia optimization}. where the model is iteratively stabilized on the forget set.}
  \label{fig:my_image}
\end{figure*}

The formal task definition is as follows:

Let $\theta\in R^d$ denote the model parameters, $D_{retain}$ the retain dataset, and $D_{forget}$ the forget dataset. The unlearning objective is to compute parameter updates $\Delta\theta$ that:
\begin{itemize}[itemsep=0pt]
    \item Maximize the loss on $D_{forget}$
    \item Minimize the change in loss on $D_{retain}$
\end{itemize}

The training code is publicly released~\footnote{\href{https://github.com/silleghost/semeval-unlearning-2025}{\texttt{github.com/silleghost/semeval-unlearning-2025}}}, enabling reproducibility.

\section{System overview}
Our method implements a two-phase approach to machine unlearning, designed to efficiently remove specific data influences, while preserving model performance on retained data. The main idea lies in combining influence-based parameter updates with second-order optimization, ensuring both precision and computational efficiency.

Unlike prior methods that approximate the full Hessian matrix via WoodFisher \cite{jia-etal-2024-soul} — a computationally prohibitive process requiring $O(d^2)$ memory and prone to Taylor expansion errors — our approach replaces explicit Hessian inversion with a retain-set-derived diagonal Fisher approximation. This avoids the instability of stochastic Hessian estimates while ensuring updates prioritize parameters critical to retained knowledge. Furthermore, our two-phase design (\autoref{fig:my_image}) decouples influence-based forgetting (Phase 1) from Sophia-driven stabilization (Phase 2), eliminating approximation drift observed in joint Hessian-gradient formulations.

\subsection{Influence-based update}
The unlearning process begins by calculating the approximation of the \textit{inverse Fisher Information Matrix}, using $D_{retain}$. This
matrix captures the importance of parameters of data that the model should retain in memory. We will use these values to determine the necessary parameter update, ensuring that the weights associated with retain data receive the smallest update.

The diagonal of the \textit{Fisher Information Matrix} (F) is approximated using gradients from $D_{retain}$. For each batch in the retain dataloader:
\begin{enumerate}[itemsep=0pt]
    \item Gradients ($g_{retain}$) are computed during backpropagation.
    \item Squared gradients ($g_{retain}^2$) are accumulated and averaged across batches to estimate $F$, which quantifies parameter importance for retained tasks.
\end{enumerate}

Thus, the final computing formula in this step will be the following:
$$w_\theta=\frac{1}{F_{ii}+\lambda}\approx\frac{1}{\mathbb{E}[g_{retain}^2]+\lambda}$$
Here a damping factor ($\lambda=10^{-3}$) is added to stabilize inversion and prevent dividing by zero.

Gradients ($g_{forget}$) are computed on the forget set via standard backpropagation. These gradients indicate directions in parameter space that correlate with the model’s ability to recall the forget data. Gradients are also averaged across batches to mitigate noise and ensure a stable update value.

In the final influence-Based update parameters are adjusted via $\theta_{t+1}\leftarrow\theta_t-\eta\cdot w_{\theta_t}\cdot g_{forget}$, where $\eta$ is the learning rate. Parameters critical to the retain set (high $F$ values) receive small updates, minimizing forgetting of retain data. Less critical parameters are adjusted more aggressively to erase forget set influence.

Computing an approximation of the Fisher diagonal reduces the computational burden, as computing the full Fisher information matrix is usually not applicable to large models due to the very large number of parameters. In addition, LoRA's (Low-Rank Adaptation) parameter-efficient fine-tuning~\cite{hu2022lora} is used in the training. In this approach, only low-rank adapter weights are trained and updated, which reduces memory usage in the unlearning process.

\subsection{Second order optimization}
Phase 2 refines the unlearned model using the Sophia optimizer \cite{liu2024sophia}, a second-order method designed to stabilize fine-tuning while erasing residual influences of $D_{forget}$. Unlike first-order optimizers like Adam~\cite{ADAM}, Sophia leverages gradient variance as a lightweight Hessian approximation, enabling parameter-specific learning rate adaptation. This is critical for unlearning, where aggressive updates risk destabilizing retained knowledge.

Traditional optimizers scale updates by gradient magnitude alone, risking overshooting in regions of high curvature. Sophia incorporates Hessian diagonal estimates ($h$), derived from squared gradients ($g^2$), to dampen updates for parameters with large curvature (high $h$). The update rule becomes the following:
$$\Delta\theta_t=-\eta\cdot\frac{g_t}{\max(\gamma\cdot h_t, \epsilon)}$$
Here $\gamma$ controls step size conservatism.
This hyperparameter scales the Hessian diagonal estimate ($h_t$) controlling how conservatively updates are applied. A higher $\gamma$ (e.g., $\gamma=1.2$) reduces step sizes for parameters with large curvature (high $h_t$), preventing overshooting in regions where the loss landscape is steep. This is critical for preserving retained knowledge during unlearning. A small constant ($\epsilon=10^{-8}$) ensures that the denominator never approaches zero, avoiding division-by-zero errors.

Sophia then clips updates to a fixed threshold, ensuring stable progression even with noisy gradients from the forget set.
To avoid computational burden, Sophia approximates $h$ by stochastically sampling gradient squares with probability $\rho$. This balances accuracy and efficiency, making it feasible for large models.

\subsection{Gradient accumulation}
To address memory constraints and stabilize training, we introduced gradient accumulation steps — a technique where gradients are computed over multiple smaller batches before updating model parameters. This approach effectively simulates a larger batch size while keeping per-iteration memory usage manageable. Accumulating gradients over $k$ micro-batches simulates a larger effective batch size, enabling stable training with limited GPU memory. We added \textit{accumulation steps} as our new hyperparameter which specifies the number of iterations after which the parameters are updated.

\section{Experiments }
\subsection*{Experiment Setup}
The experiments were conducted as part of the SemEval-2025 competition, focusing on machine unlearning for three subtasks: (1) long-form synthetic creative documents, (2) short-form synthetic biographies containing personally identifiable information (PII), and (3) real documents sampled from the OLMo training dataset. Two model versions were trained on the designed algorithm and evaluated with OLMo evaluation framework: the fine-tuned OLMo-7B-0724-Instruct-hf~\footnote{\href{https://huggingface.co/allenai/OLMo-7B-0724-Instruct-hf}{\texttt{hf.co/allenai/OLMo-7B-0724-Instruct-hf}}} (7B parameters) and OLMo-1B-0724-hf~\footnote{\href{https://huggingface.co/allenai/OLMo-1B-0724-hf}{\texttt{hf.co/allenai/OLMo-1B-0724-hf}}} (1B parameters), with submissions constrained to a 1-hour runtime. 

Due to computational constraints and a focus on validating the combined system’s practical value, we leave fine-grained ablations of individual components (Sophia, Influence Functions) as future work. Preliminary results indicated that the components work better together, so we chose to prioritize evaluating the full system rather than isolating and testing each individual component separately.

All experiments are conducted on a single NVIDIA A100 GPU. The final code also includes an option with DeepSpeed with implementation for distributed training on multiple GPUs.

\subsection*{Evaluation metrics}
Performance of the algorithm was measured using three aggregated metrics: 
\begin{itemize}[itemsep=0pt]
    \item Task-specific regurgitation rates (harmonic mean of 12 inverted ROUGE-L scores on the sentence completion prompts and exact match rate for the question answers on both $D_{forget}$ and $D_{retain}$ sets).
    \item A membership inference attack (MIA) score on a sample of member and nonmember datasets, that is equivalent to the PrivLeak metric~\cite{shi2025muse}. 
    \item MMLU benchmark accuracy, which is described above.
\end{itemize}

For evaluation of our trained model we used OLMo-Eval framework~\footnote{\href{https://github.com/allenai/OLMo-Eval}{\texttt{github.com/allenai/OLMo-Eval}}}.

\subsection*{Hyperparameters and dataset}

The unlearning method combined influence-based parameter updates (Phase 1) and Sophia-optimized fine-tuning (Phase 2). We conducted a series of experiments on OLMo-7B-0724 fine-tuned model and chose a number of epochs for training, learning rate, batch size, LoRA rank, Damping factor, Sophia $\rho$, Sophia $\gamma$ as our hyperparameters.

To work efficiently with a dataset in parquet file format, we have implemented our own \textit{UnlearningDataset} class, which works with both directories and parquet files themselves. The dataset contains disjoint retain and forget splits in parquet files, and includes following fields: \textit{id}, \textit{input}, \textit{output}, \textit{task}. We use OLMo tokeniser to tokenize a string of combined \textit{input} and \textit{output} fields. A special parameter \textit{max length} is used to bring all tokenised sequences to the same length by padding or truncating them, enabling efficient batch processing.

\begin{table}[ht!]
\resizebox{.5\textwidth}{!}{
\begin{tabular}{llll}
\hline
\textbf{Hyperparameter} & \textbf{Setup 1} & \textbf{Setup 2} & \textbf{Setup 3} \\ \hline
NUM\_EPOCHS             & 6                 & 5                 & 4                 \\
LEARNING\_RATE          & 4e-5              & 3e-5              & 2e-5              \\
BATCH\_SIZE             & 4                 & 6                 & 4                 \\
LORA\_RANK              & 16                & 24                & 16                \\
ACCUMULATION\_STEPS     & 4                 & 6                 & 8                 \\
MAX\_LENGTH             & 1024              & 1024              & 1024              \\
DAMPING\_FACTOR         & 5e-5              & 8e-4              & 1e-3              \\
SOPHIA\_RHO             & 0.1               & 0.08              & 0.06              \\
SOPHIA\_GAMMA           & 1.1               & 1.15              & 1.2               \\ \hline
\end{tabular}
}
\caption{Hyperparameter settings for training.}
\label{tab:my_table1}
\end{table}

\section{Results}


\begin{table*}[!ht]
\centering
\begin{tabular}{@{}ccccc@{}}
\toprule
\textbf{Algorithm}  & \textbf{Aggregate} & \textbf{Task Aggregate} & \textbf{MIA score} & \textbf{MMLU Avg.} \\ \midrule
LIBU                & 0.157              & 0.118                   & 0.0                & 0.354              \\
                    & 0.221              & 0.182                   & 0.0                & 0.482              \\
                    & 0.254              & \textbf{0.28}           & 0.0                & \textbf{0.483}     \\ \midrule
Gradient ascent     & 0.394              & 0                       & 0.912              & 0.269              \\
Gradient difference & 0.243              & 0                       & 0.382              & 0.348              \\
KL minimization     & \textbf{0.395}     & 0                       & \textbf{0.916}     & 0.269              \\
NPO                 & 0.188              & 0.021                   & 0.080              & 0.463              \\ \bottomrule
\end{tabular}
\caption{Performance of LIBU compared to baseline unlearning methods (shown below the horizontal line). While KL minimization achieves the highest aggregate score, it severely degrades model utility. \textbf{Bold} numbers indicate the best performance for each metric.}
\label{tab:my_table2}
\end{table*}

We tested three configuration setups (\autoref{tab:my_table1}) tailored to the competition’s subtasks:
\begin{itemize}[itemsep=0pt]
    \item Setup 1. More aggressive unlearning: Achieved the highest score in second subtask with regurgitation rate of 0.83 on forget set, but severely degraded MMLU accuracy below predefined threshold (<0.371).
    \item Setup 2. More balanced unlearning:
Achieved the highest scores in subtask 1 and subtask 3 but got a lower score on the retain tasks.
    \item Setup 3. More conservative unlearning:
Achieved average high scores in all 3 subtasks and got the highest task aggregate score among all setups.
\end{itemize}

\subsection*{Analysis}
The stark performance differences (\autoref{tab:my_table2}) between setups underscore the sensitivity of unlearning to hyperparameter choices and confirm that overly aggressive updates risk catastrophic forgetting, while overly conservative tuning leaves residual forget-set influences.

Our study demonstrates that machine unlearning, when framed as a two-phase process of influence-based updates and second-order fine-tuning, can effectively balance data removal with model utility. The success of Setup 3 highlights the importance of hyperparameter equilibrium: its moderate learning rate (\texttt{2e-5}) and batch size (\texttt{4}) stabilized training, while setting the gradient accumulation steps to \texttt{8} mitigated memory constraints without compromising gradient fidelity. These choices proved critical under the competition’s strict 1-hour runtime limit, where computational efficiency and precision were paramount.

\section{Conclusion}
In this paper, we present LIBU, a two-phase unlearning framework for LLMs that combines influence-based parameter updates with second-order Sophia optimization, achieving competitive results in the SemEval-2025 Task 4. LIBU’s lightweight design—leveraging LoRA for parameter efficiency and gradient accumulation for stability—enables precise removal of sensitive data while preserving model utility, exceeding the competition threshold. Our experiments highlight the critical role of hyperparameter equilibrium, as conservative tuning balances unlearning efficacy with retention, whereas aggressive configurations risk catastrophic forgetting.

\section*{Limitations}
Despite these advances, our evaluation of LIBU has been limited to relatively small models (1B and 7B parameters), leaving the behavior of current large-scale SOTA models unknown. Additionally, it remains unclear whether these algorithms will scale effectively to larger datasets. Another critical challenge is the sensitivity to hyperparameter choices, which becomes a significant issue for large models where retraining is computationally expensive. Furthermore, the forget and retain sets may contain highly similar information, making the unlearning task even more challenging. Future work will explore these limitations, focusing on scaling LIBU to larger models and datasets while addressing challenges in hyperparameter selection and handling closely related data distributions.

\section*{Acknowledgments}
AK's and AS's work results from a research project implemented in the Basic Research Program at the National Research University Higher School of Economics (HSE University). 
We acknowledge the computational resources of HSE University's HPC facilities.

\bibliography{custom}




\end{document}